# SYSTEM IDENTIFICATION AND MODELING FOR INTERACTING AND NON-INTERACTING TANK SYSTEMS USING INTELLIGENT TECHNIQUES

Bhuvaneswari N S[1], Praveena R[2], Divya R[3]

Electronics and Instrumentation Engineering Department,
Easwari Engineering College, Chennai, Tamilnadu, India

[1]bhuvaneswarins@rediff.com, [2]praveenarmn26@gmail.com, [3]awesomerads@gmail.com

## ABSTRACT

*System identification from the experimental data plays a vital role for model based controller design. Derivation of process model from first principles is often difficult due to its complexity. The first stage in the development of any control and monitoring system is the identification and modeling of the system. Each model is developed within the context of a specific control problem. Thus, the need for a general system identification framework is warranted. The proposed framework should be able to adapt and emphasize different properties based on the control objective and the nature of the behavior of the system. Therefore, system identification has been a valuable tool in identifying the model of the system based on the input and output data for the design of the controller. The present work is concerned with the identification of transfer function models using statistical model identification, process reaction curve method, ARX model, genetic algorithm and modeling using neural network and fuzzy logic for interacting and non interacting tank process. The identification technique and modeling used is prone to parameter change & disturbance. The proposed methods are used for identifying the mathematical model and intelligent model of interacting and non interacting process from the real time experimental data.*

## 1. INTRODUCTION

The process of constructing models from experimental data is called system identification. System identification involves building a mathematical model of a dynamic system based on set of measured stimulus and response samples. It is a process of acquiring, formatting, processing and identifying mathematical models based on raw data from the real-world system. Once the mathematical model is chosen, it can be characterized in terms of suitable descriptions such as transfer function, impulse response or power series expansions and that can be used for controller design. Tri Chandra S.Wibowo.et.al [1] have done System Identification of an Interacting Series Process for Real-Time Model Predictive Control. This paper aims at identifying a linear time-invariant (LTI) with lumped parameters state space model of the gaseous pilot plant which has a typical structure of interacting series process and the model has been developed around an operating point. Edward P. Gatzke.et.al [2] have done work on Model based control of a four-tank system. In this paper, the authors used sub space process modeling and hence applicable to particular operating point.Nithya.et.al [3] have done work on model based controller design for a spherical tank process in real time. They have proposed tangent and point of inflection methods for estimating FOPTD model parameters. The major disadvantage of all these methods is the difficulty in locating the point of inflection in practice and may not be accurate. Gatzke *et al* [2] perform the parametric identification process of a quadruple tank using subspace system identification method. Such a system has series structure with recycles and the input signals used are the pseudo-random binary sequence (PRBS)..The identification process is carried out without taking into account the prior knowledge of process, and no assumption are made about the state relationships or number of process states. Weyer [5] presents the empirical modeling of water level in an irrigation channel using system identification technique with taking into account the prior physical information of the system. The identified process is a kind of interacting series process, however, the model only has a single output variable.

To accurately control a system, it is beneficial to first develop a model of the system. The main objective for the modeling task is to obtain a good and reliable tool for analysis and control system development. A good model can be used in off-line controller design and implementation of new advanced control schemes. In some applications, such as in an industrial sewing machine[6] , it may be time consuming or dangerous to tune controllers directly on the machinery. In such cases, an accurate model must be used off-line for the tuning and verification of the controller.

In the present work, The system identification of Interacting and Non-interacting tank systems are found using genetic algorithm which is working out for full region and the results are compared with Process Reaction curve method, ARX model, and Statistical model of Identification. Also, we propose a method to obtain an accurate nonlinear system model based on neural networks (NNs) and Fuzzy logic. Modeling techniques based on NNs and Fuzzy logic have proven to be quite useful for building good quality models from measured data. The paper is organized as follows. In Section-II, the process description is discussed. In section-III, the process reaction curve method for the problem considered is discussed. In section-IV, the statistical model of system identification implementation is discussed. In section-V, the ARX model development is discussed.In section –VI, the Genetic Algorithm for System Identification is discussed. In the last section, the Neuro and Fuzzy models and finally, the paper is concluded.

## 2. PROCESS DESCRIPTION

In the present work, the real time interacting and non-interacting fabricated system was used for collecting the input, output data. The setup consists of supply tank, pump for water circulation, rotameter for flow measurement, transparent tanks with graduated scales, which can be connected, in interacting and non-interacting mode. The components are assembled on frame to form tabletop mounting. The set up is designed to study dynamic response of single and multi capacity processes when connected in interacting and non-interacting mode. It is combined to study Single capacity process, Non-interacting process and Interacting process. The experimental set up is shown in Figure 1. The specifications are tabulated in Table 1. The schematic diagrams of Interacting and Non-interacting systems are shown in the Figure 2 & 3.

Table 1. Specifications of the set up

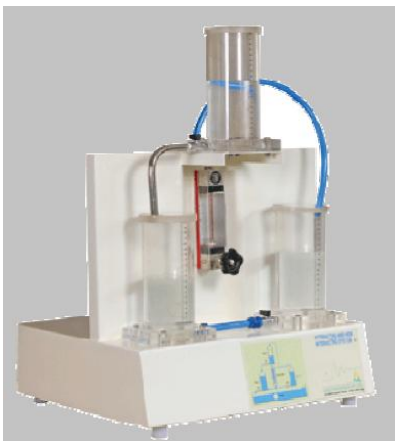

| Components | Details |
|---|---|
| Rotameter | 10-100 LPH |
| Process tank | Acrylic, Cylindrical, Inside Diameter 92mm With graduated scale in mm. (3 Nos) |
| Supply tank | SS304 |
| Pump | Fractional horse power, type submersible |
| Overall dimensions | 550Wx475Dx520H mm |

Fig.1.Experimental Set up

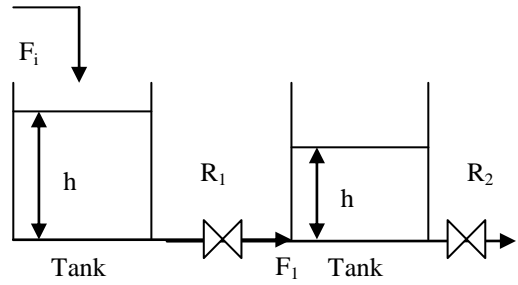

Figure 2. Schematic diagram of Interacting process

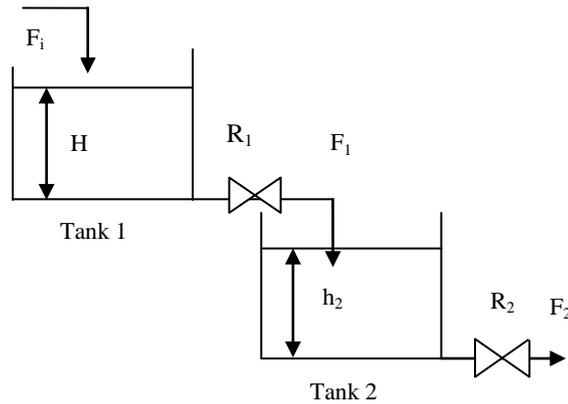

Figure 3. Schematic diagram of Interacting process

The dynamic response for the interacting system is given by

$$y1 = k_p*(20 - \exp(-t/t1*t2(\exp(-(t/(t1+t2+AR)))) \quad (1)$$

Table.2 & 3 show the steady state and dynamic data generated from the interacting process.

Table.2. Steady state data for Interacting system

| Flow (lph) | $h_2$ (mm) | $h_1$ (mm) |
|---|---|---|
| 20 | 35 | 43 |
| 30 | 43 | 60 |
| 40 | 55 | 85 |

Table.3. Dynamic data for Interacting system for $F_{in}$ = 20 lph

| Time (s) | Level of tank (mm) | Observed (mm) |
|---|---|---|
| 0 | 1 | 0 |
| 15 | 2 | 5 |
| 30 | 7 | 10 |
| 40 | 12 | 14 |
| 50 | 16 | 19 |
| 60 | 21 | 22 |
| 70 | 24 | 25 |
| 80 | 27 | 28 |
| 90 | 32 | 30 |
| 95 | 35 | 33 |
| 100 | 36 | 34 |

The dynamic response for the non-interacting process is given by

$$y(t)=K'_p\left[1+\frac{1}{\tau_{p2}-\tau_{p1}}\left(\tau_{p1}e^{-t/\tau_{p1}}-\tau_{p2}e^{-t/\tau_{p2}}\right)\right] \quad (2)$$

Table.4&5 show the steady state and dynamic data collected form the non-interacting process.

Table.4.Steady state data for Non-interacting system

| Flow (lph) | $h_1$ (mm) | $h_2$ (mm) |
|---|---|---|
| 30 | 90 | 140 |
| 40 | 110 | 95 |
| 50 | 125 | 75 |

Table.5.Dynamic data for Non-interacting System for $F_{in}$ = 30lph

| Time(t) (sec) | Level of tank (T2)(mm) | Level of tank (T1) (mm) |
|---|---|---|
| 0 | 0 | 0 |
| 15 | 10 | 20 |
| 30 | 18 | 28 |
| 45 | 26 | 42 |
| 60 | 35 | 50 |
| 75 | 45 | 60 |
| 90 | 55 | 66 |
| 105 | 65 | 74 |
| 120 | 78 | 79 |
| 135 | 90 | 85 |
| 150 | 102 | 89 |
| 165 | 115 | 91 |
| 180 | 126 | 91 |
| 195 | 139 | 91 |
| 210 | 145 | 91 |

After collecting the data, the next step is to obtain the transfer function model using process Reaction curve method that is discussed in the next section.

## 3. PROCESS REACTION CURVE METHOD

In this section, the transfer function model using process reaction curve method is discussed for which the input and output data are generated from the real time interacting and non-interacting system. This method is kept as the base model for comparing other methods of system identification.

### 3.1 Transfer function model for interacting system

The steady state and dynamic data obtained for Interacting process are tabulated in the Tables 2&3 from which the steady state graph and process reaction curve are plotted.

From the slope of the graphs, the following parameters are measured
Resistance $R_1 = dh_1/dt = 1.5$ohm
Resistance $R_2 = dh_2/dt = 1.7$ohm
Diameter of tank $T_2 = 92$cm
Diameter of tank $T_1 = 92$cm
Initial flow = 20 lph

Step amplitude 40 − 20 = 20 m$^3$/sec
Time constant $\tau_1 = A_1R_1 = 0.9$
Time constant $\tau_2 = A_2R_2 = 1.02$

Using the above parameters, the transfer function for the interacting system obtained is

$$TF = \frac{1.7}{0.918s^2 + 1.93s + 1} \quad (3)$$

## 3.2 Transfer function model for Non-interacting system

The steady state and dynamic data obtained for Interacting are tabulated in the Tables 4&5 from which the steady state graph and process reaction curve are plotted.

From the slope of the graphs, the following parameters are measured

$R_1 = 4.6$ ohms
$R_2 = 1.5$ ohms
Diameter of tank1 and tank2 = 92 cm
Step Amplitude = 60 − 30 = 30 m$^3$/sec
Time constant = $\tau_1 = A_1R_1 = 0.9954$
$\tau_2 = A_2R_2 = 3.049$

$$TF = \frac{R_2}{[(A_1R_1)s+1][(A_2R_2)s+1]} \quad (4)$$

$$= \frac{4.6}{[(0.9954s+1)(3.049s+1)]} \quad (5)$$

Thus the transfer functions are obtained with process reaction curve methods. But the major disadvantage of this method is difficulty in locating the point of inflection in practice and may not be accurate. Gatzke *et al* [4]. Hence the other methods are suggested here and the Statistical model identification method is discussed in the next section.

## 4. STATISTICAL MODEL OF SYSTEM IDENTIFICATION

Statistical model identification methods provide more flexible approaches to identification that relax the limits to model structure experimental design. In addition, the statistical method uses all data and not just a few points from the response, which provide better parameter estimates from noisy process data. The input and output data are taken from the process for every sampling instance. Using these data, the Z matrix and U matrix are formed which is the basis for calculating the parameters of the system.

### 4.1. Formation of Z Matrix

Z matrix is formed using the output data. Each data is subtracted from the first output data. The first non-zero values from the difference output are considered as the Z matrix.

### 4.2. Formation of U Matrix

U matrix consists of four columns which involves both the input and output data. The first and second column is just one shift from the Z matrix. The third and fourth column is formed using the input data each data is subtracted from the first input data. The first non-zero values from the difference input are considered as the second column of the U matrix.

### 4.3. Calculation of System Parameters of Second order System Identification by Least Squares Regression

The second order model structure is
$Y(k) = a_1 y(k-1) + a_2 y(k-2) + b_1 u(k-1) + b_2 u(k-2)$ (6)

Where $a_1, a_2, b_1, b_2$ are model parameters
If the system is tested with the input signal
$u(k), k \in \{1,2,.....N\}$
and the measured corresponding output is
$y(k), k \in \{1,2,....N\}$
and define $\theta=\{\theta_1\ \theta_2\ \theta_3\ \theta_4\}=\{a_1\ a_2\ a_3\ a_4\}$, then
$\varphi(k)=[y(k-1)\ y(k-2)\ u(k-1)\ u(k-2)]$

Then, the model structure in matrix notation to be $Y=\emptyset.\theta$
By using the vector least square regression, calculated using the equation defined below, while the estimated parameter vector of $\theta$

$$\hat{\theta}=(\emptyset'\emptyset)^{-1}\emptyset'Y \qquad (7)$$

Using the above method, the parameters obtained for the interacting system are,
$a_1 = -0.8215 \qquad a_2 = 1.814$
$b_1 = 0 \qquad\qquad b_2 = 0.0129$

Using these parameters, the transfer function obtained is,
$$G(z) = \frac{0.0129}{Z^2 - 1.814z + 0.8215}$$

Using the above method, the parameters obtained for the Non-nteracting system are,
$a_1 = -0.7973 \qquad a_2 = 1.7817$
$b_1 = 0 \qquad\qquad b_2 = 0.0707$

On these, the transfer function obtained is,
$$TF = \frac{0.0707}{Z^2 - 1.7817Z + 0.7973}$$

Thus the transfer functions are found by using the SMI method. In the next section the ARX model of system identification is discussed.

## 5. ARX MODEL OF SYSTEM IDENTIFICATION FOR INTERACTING AND NON-INTERACTING PROCESS

ARX means "Auto regressive eXternal input". It is considered as black box system which can be viewed in terms of input, output and transfer characteristics without the knowledge of its internal working. To assess the data and the degree of difficulty in identifying a model, first estimate the simplest, discrete-time model to get a relationship between u(t) and y(t), the ARX model which is discussed here.

The ARX model is a linear difference equation that relates the input u(t) to the output y(t) as follows:

$y(t) + a_1 y(t-1) + ... + a_{n_a} y(t-n_a) = b_1 u(t-1) + ... + b_{n_b} u(t - n_k - n_b) + e(t)$

Since the white-noise term e(t) here enters as a direct error in the difference equation, the above equation is often called as an equation error model (structure). The adjustable parameters are in this case,

$\theta = [a_1\ a_2 ... a_{n_a}\ b_1 ... b_{n_b}]^T$

### 5.1. Model for Interacting Process

orders = [ 2 1 1 ]
z = [ y ,u ] ;
m= arx ( z ,orders)

Discrete-time IDPOLY model: A(q)y(t)=B(q)u(t)+ e(t)

$A(q) = 1 – 1.824 q^{-1} + 0.835 q^{-2}$
$B(q) = 0.0182 q^{-1}$

The parameters obtained for Interacting process are
$a_1 = -0.7991$ and $a_2 = 1.783$
$b_1 = 0$ and $b_2 = 0.0768$

The transfer function obtained for interacting process is

$$TF = \frac{0.0182}{Z^2 - 1.824Z + 0.835} \qquad (8)$$

## 5.2. Model for Non-Interacting Process

orders = [ 2 1 1 ]
z = [ y ,u ] ;
m= arx ( z ,orders)

Discrete-time IDPOLY model: $A(q)y(t) = B(q)u(t) + e(t)$
$A(q) = 1 - 1.782 q^{-1} + 0.7973 q^{-2}$
$B(q) = 0.07068 q^{-1}$

The parameter obtained,
$a_1 = -0.7991$ and $a_2 = 1.783$
$b_1 = 0$ and $b_2 = 0.0768$

The transfer function obtained for non-interacting process in matlab,

$$TF = \frac{0.0768}{Z^2 - 1.783z + 0.7991} \qquad (9)$$

Thus the transfer functions are developed using ARX model. In the next section, the Genetic Algorithm for identification of the considered process is developed.

## 6. GENETIC ALGORITHM FOR INTERACTING AND NON-INTERACTING PROCESS

A genetic algorithm (GA) is a method for solving both constrained and unconstrained optimization problems based on a natural selection process that mimics biological evolution. The algorithm repeatedly modifies a population of individual solutions. At each step, the genetic algorithm randomly selects individuals from the current population and uses them as parents to produce the children for the next generation. Over successive generations, the population "evolves" toward an optimal solution.

In this work, GA tool is used for finding the parameters by comparing with realtime input, output data. The steps for designing GA process are given below.

### 6.1. Steps for designing GA for Interacting and Non-interacting Process System Identification

1. Open the GA tool
2. Call the objective function
   a. In the objective function, define the parameters to be found as function arguments.
   b. Write the dynamic response expression for the interacting and Non-interacting systems in terms of the parameters to be found.
   c. Call the input, output data generated.
   d. Running GA will randomly substitute the parameters value and the output is found from the dynamic response expression.
   e. Compare and generate error between this output from the expression and the dataset.
   f. Square and sum up this error till the time, until the steady state value .

g. GA will continuously run until this sum square error is zero.
   h. Once, the sum square error is zero, the values obtained will be optimum.
3. Initialize and enter the range for the parameters, so that it is easy for the tool to check within the range and take minimum time to find the optimum parameters..
4. Enter the number of population
5. Enter the stopping criteria parameters.
6. Select all the plot functions, so that the error, sum square error and iterations all are visualized easily.

Table 6 shows the parameters set in GA tool. Running GA tool will end towards optimum solutions. The parameters given by the GA tool is validated with real-time dataset.

Table 6. Parameters used in GA tool

| Objective function | Integral Square error (ISE) |
|---|---|
| Initial Ranges for the parameters | $a_1$ – 0 to 1<br>$b_1$ – 0 to 1<br>$a_2$ – 0 to 5<br>$b_2$ – 0 to 1 |
| Initial population | 10 |
| Population size | 24 |
| Cross over | Multipoint crossover |
| Stopping Criteria | ISE = 0 |

# 7. Development of Neuro and Fuzzy model for Interacting and Non- interacting process

The modeling of interacting and non-interacting process in process industries is an extremely difficult task. This is because the dynamics of the process are non-linear and time varying. Hence it is necessary to employ universal approximators like neural networks and Fuzzy logic which perform mapping of the input-output data set. This mapping is conformal in nature. Hence the neural network and Fuzzy logic behave like a "BLACK BOX" which can be used to approximate any non-linear, time varying process.

### 7.1. Design of neuro modeling

The major steps involved in the design of modeling using neural networks are as follows:

a) **Identification of the architecture of the Neural Network:** Whenever the input-target patterns are known, the Multi Layer Perceptron (MLP) is the best choice as it does the exact
input-output mapping. Hence, the type of neural network best suited for interacting and non-interacting process is the MLP having three layers i.e. Input layer, hidden layer, and output layer. The number of nodes in the hidden layer is determined to be five, by trial and error after considering the training error after 6000 epochs. The activation function for the hidden layer is chosen to be the binary sigmoid whose output varies from 0 to 1 and for the output layer, it is linear. The inputs applied to the neural network are inflow($F_{in}$) and previous output(h(k-1)).

b) **Obtaining the training data:** The training data are generated from the open loop response of the system for various step changes in inflow. Sample of data used for training Non-interacting process and Interacting process are shown in the Tables 7 & 8. Care is taken to ensure that the training data set is a representative of the type of input and output patterns encountered so that the model could be effectively designed.

Table.7.Sample of data used for Non-interacting process    Table.8.Sample of data used for Interacting process

| Inflow in Lph | H(t-1) | H(t) |
|---|---|---|
| 30 | 0 | 0.0006 |
| 30 | 0.0006 | 0.0154 |
| 30 | 0.0154 | 0.3673 |
| 30 | 0.3673 | 3.3981 |
| 30 | 3.3981 | 12.2047 |
| 30 | 12.2047 | 27.7299 |
| 30 | 27.7299 | 48.0532 |
| 30 | 48.0532 | 68.3283 |
| 30 | 68.3283 | 87.1988 |
| 30 | 87.1988 | 103.4765 |
| 30 | 103.4765 | 116.6009 |
| 30 | 116.6009 | 126.3754 |
| 30 | 126.3754 | 132.9183 |
| 30 | 132.9183 | 136.6447 |
| 30 | 136.6447 | 137.4902 |
| 30 | 137.4902 | 137.8086 |

| Inflow in Lph | H(t-1) | H(t) |
|---|---|---|
| 20 | 0 | 0.0011 |
| 20 | 0 | 0.0278 |
| 20 | 0.0011 | 0.553 |
| 20 | 0.0278 | 2.0985 |
| 20 | 0.553 | 4.2443 |
| 20 | 2.0985 | 6.7081 |
| 20 | 4.2443 | 9.2934 |
| 20 | 6.7081 | 11.867 |
| 20 | 9.2934 | 14.3416 |
| 20 | 11.867 | 16.6635 |
| 20 | 14.3416 | 18.8029 |
| 20 | 16.6635 | 20.7465 |
| 20 | 18.8029 | 22.4926 |
| 20 | 20.7465 | 24.0473 |
| 20 | 22.4926 | 25.421 |
| 20 | 24.0473 | 26.6272 |

c) **Training:** The most important step involved in the design of the neural network model is the training of the Neural Network. The architecture used for the training of the neuromodel is shown in the Figure 4. The training is accomplished by the use of Neural Network Toolbox (NNTOOL) available in MATLAB. The training patterns of inflow and previous height are concatenated into an n×2 matrix where n is the number of training patterns. The output (current height) is presented to NNTOOL as a column vector. There are many algorithms present in NNTOOL for training. Among these, a gradient descent algorithm with momentum factor included (TRAINGDM), is used for training. The stopping criterion specified is 0.25, i.e. the training is stopped when the Root mean square error (RMSE) between the network outputs and the targets is lesser than or equal to 0.25. The learning rate is fixed at 0.5. The number of training epochs is fixed uniformly at 6000. The number of hidden nodes are changed from 2 to 6.For 5 hidden nodes the error is found to be minimum as shown in the Table 9 and it is used in training the network. Some of the patterns are also used to test the network in order to prevent over fitting of the training data. Finally the values of the weights obtained after training is used for the feed forward implementation.

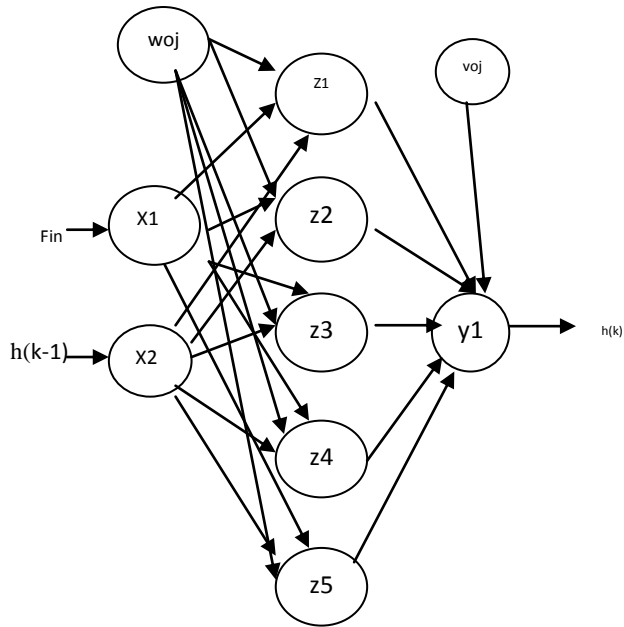

**Figure 4. Architecture of Neuro Model**

**Table 9   Selection of number of nodes in hidden layer**

| No. of Hidden Nodes | RMSE after 6000 epochs |
|---|---|
| 2 | 5.6 |
| 3 | 3.2 |
| 4 | 1.7 |
| 5 | 0.15 |
| 6 | 0.83 |

7.2. Design of Fuzzy Modeling

The design of fuzzy control system consists of several steps. First the variables for the fuzzy model are selected. The universe of discourse for all the variables involved are then set. Here the fuzzy model is designed with two input variables inflow($F_{in}$) and Previous output(h(k-1)) and one output variable, current output(h(k)). The universe of discourse for these parameters are 0 to 300 lph, 0 to 60 cm and 0 to 60 respectively. The membership functions of inflow, previous output and the current output are generated in the Matlab Fuzzy toolbox and the Fuzzy design was done.  The results with Neuro and Fuzzy model for Interacting and Non-interacting process are shown in the figures 5 & 6.

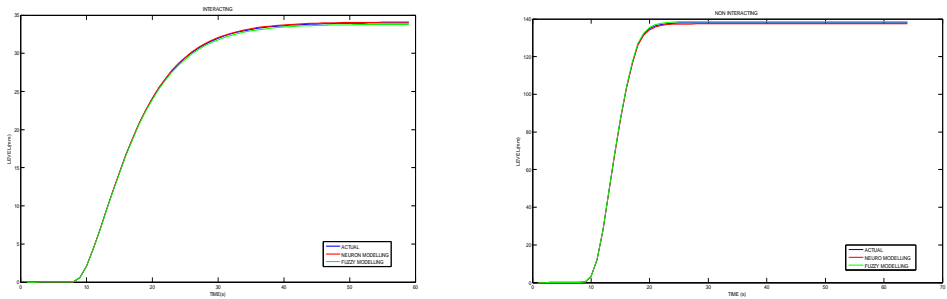

Figs.5 & 6. Dynamic response of Interacting and Non-interacting systems with Neuro and Fuzzy modeling

# 8. RESULTS AND CONCLUSIONS

The parameters obtained by various methods for interacting and non-interacting process are tabulated in the Table 10 & 11. These values are substituted in the model and validated for the input and are shown in the Figure 4 & 5.

The results obtained from the process reaction curve identification method are compared with statistical method, ARX model and genetic algorithm method and System identification using GA is found to be one of the simple methods for system identification without need for much of calculations.

Table 10. Comparison of Parameters obtained for Interacting process

| Methodology | $K_{p1}$ | $K_{p2}$ | $\tau_1$ | $\tau_2$ | Transfer function |
|---|---|---|---|---|---|
| PRC method | 1.7 | - | 1.02 | 0.9 | $\frac{1.7}{0.918s^2+1.93s+1}$ |
| Genetic Algorithm | 2.35 | 0.7 | 1.02 | 0.9 | $\frac{1.65}{0.918S^2+1.93s+1}$ |
| Parameters | $a_1$ | $a_2$ | $b_1$ | $b_2$ | Transfer function |
| Statistical method | 0.8215 | 1.814 | 0 | 0.0129 | $\frac{0.0129}{z^2-1.814z+0.8215}$ |
| ARX method | 0.835 | 1.824 | 0 | 0.0182 | $\frac{0.0182}{z^2-1.824z+0.835}$ |

Table 11. Comparison of parameters obtained for Non-interacting process

| Methodology | $K_{p1}$ | $K_{p2}$ | $\tau_1$ | $\tau_2$ | Transfer function |
|---|---|---|---|---|---|
| PRC method | 4.6 | - | 3.049 | 0.9954 | $\frac{4.6}{3.035s^2+4.045s+1}$ |
| genetic algorithm | 2.98 | 1.53 | 3.049 | 0.9954 | $\frac{4.55}{3.035S^2+4.045S+1}$ |
| parameters | $a_1$ | $a_2$ | $b_1$ | $b_2$ | Transfer function |
| statistical method | 0.7973 | 1.7817 | 0 | 0.0707 | $\frac{0.0707}{z^2-1.7817z+0.7973}$ |
| arx method | 0.7991 | 1.7882 | 0 | 0.0768 | $\frac{0.0768}{z^2-1.783z+0.7991}$ |

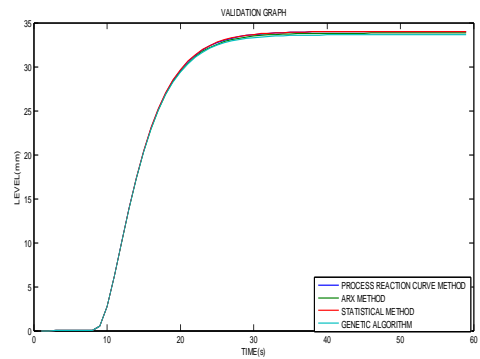

Figure 4. Validation graph for Inteacting process for the input of 20 lph.

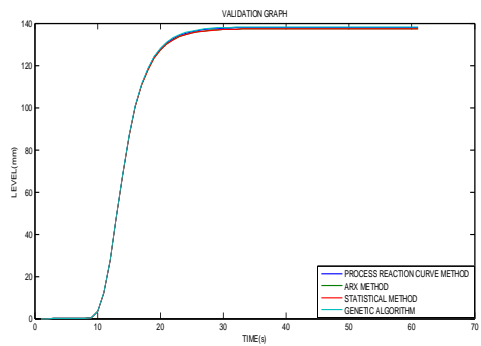

Figure.5. Validation graph for Non-Inteacting process for the input of 30 lph.

Authors

1.Dr.N.S.BHUVANESWARI obtained her B.E degree in Electronics & Instrumentation 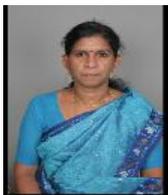 branch from Annamalai University, M.S by Research from MIT,Anna University, Chennai and Ph.D from Anna University,Chennai. She has 22 years of teaching experience in various Engineering colleges. Presently she is working as Prof & Head, EIE Department,Easwari Engineering College,Chennai.She has published 5 Journal papers in various International Journals.She has presented several technical papers in National & International Conferences across the Country.She has organized many national level workshops, Conferences,Seminars and Faculty Development Programmes.She has Chaired the paper presentation session in national and international conferences.At present she is guiding two Ph.D scholars from Anna University.Her fields of interest are Adaptive, Optimal and various Intelligent controls.

2. 2. R.PRAVEENA obtained her B.E degree in Instrumentation & Control branch from Arulmigu Kalasalingam college of Engineering. After qualifying in GATE exam she 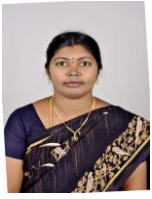 obtained her M.E in Control & Instrumentation branch from College Of Engineering, Guindy, Anna University, Chennai. She has 4 years of teaching experience in various Engineering colleges. Presently she is working as Assistant Prof, EIE Department, Easwari Engineering College, Chennai.

3. R.DIVYA completed her B.E degree in Electronics & Instrumentation from Easwari 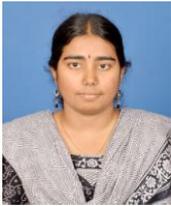 Engineering College, affiliated to Anna University, Chennai. She has been a member of ISA society for two years. She holds good academic record. She has been a Joint Secretary for Symposium INSTRUBLITZ. She has attended various workshops such us Beagle board of Texas instruments at Anna University, Chennai , Robotics and automation and Evolution of embedded systems by SPIRO limited. She has Presented Papers on Magnetic resonance imaging technique and Neural networks & face recognition.